\pdfoutput=1

\documentclass[11pt]{article}

\usepackage[]{acl}
\usepackage{booktabs} 
\usepackage{adjustbox}
\usepackage{colortbl} 
\usepackage{booktabs}
\usepackage{tabularx}
\usepackage{nicematrix}
\usepackage{times}
\usepackage{latexsym}
\usepackage{multirow}
\usepackage{amssymb}
\usepackage{pgfplots}  
\pgfplotsset{compat=1.17}  
\usepackage{pgfplotstable}  
\usepackage{siunitx}  
\usepackage{subfiles}
\usepackage[T1]{fontenc}

\usepackage[utf8]{inputenc}

\usepackage{microtype}

\usepackage{inconsolata}

\usepackage{graphicx}

\title{TwT: Thinking without Tokens by Habitual Reasoning Distillation\\ with
Multi-Teachers' Guidance}

\author{
Jingxian Xu\textsuperscript{\rm 1}\thanks{\ \ Work during internship at Microsoft.},
Mengyu Zhou\textsuperscript{\rm 4}\thanks{\ \ Corresponding author (mengyu.chou@gmail.com).},
Weichang Liu\textsuperscript{\rm 2}\footnotemark[1],
Hanbing Liu\textsuperscript{\rm 3}\footnotemark[1],
\\
\textbf{
Shi Han\textsuperscript{\rm 4},
Dongmei Zhang\textsuperscript{\rm 4}
}
\\
\textsuperscript{\rm 1}Nankai University 
\textsuperscript{\rm 2}BeijingJiaoTong University \\
\textsuperscript{\rm 3}Tsinghua  University
\textsuperscript{\rm 4}Microsoft Research \\
}

\begin{document}
\maketitle
\begin{abstract}
Large Language Models (LLMs) have made significant strides in problem-solving by incorporating reasoning processes. However, this enhanced reasoning capability results in an increased number of output tokens during inference, leading to higher computational costs. To address this challenge, we propose \textit{\textbf{TwT}} (Thinking without Tokens), a method that reduces inference-time costs through habitual reasoning distillation with multi-teachers' guidance, while maintaining high performance. Our approach introduces a Habitual Reasoning Distillation method, which internalizes explicit reasoning into the model’s habitual behavior through a Teacher-Guided compression strategy inspired by human cognition. Additionally, we propose Dual-Criteria Rejection Sampling (DCRS), a technique that generates a high-quality and diverse distillation dataset using multiple teacher models, making our method suitable for unsupervised scenarios. Experimental results demonstrate that TwT effectively reduces inference costs while preserving superior performance, achieving up to a 13.6\% improvement in accuracy with fewer output tokens compared to other distillation methods, offering a highly practical solution for efficient LLM deployment.
\end{abstract}

\section{Introduction}
Large Language Models (LLMs) have demonstrated remarkable improvements in problem-solving by incorporating reasoning process  \cite{brown2020language,wei2022chain,chowdhery2023palm,yao2024tree}. It enhances the reasoning capability of LLMs by breaking down complex tasks into intermediate steps, leading to better performance. However, reasoning capability comes at a significant cost: the reasoning process substantially increases the number of output tokens during inference, resulting in higher inference-time computational costs \cite{snell2024scaling,wu2024inference}. In practical deployments, computational resources and budgets are often constrained, making the reduction of inference-time computational costs a pressing issue that requires effective solutions \cite{zheng2022alpa,chung2024scaling}.
\begin{figure}[t] 
	\centering 
	\includegraphics[width=0.42\textwidth]{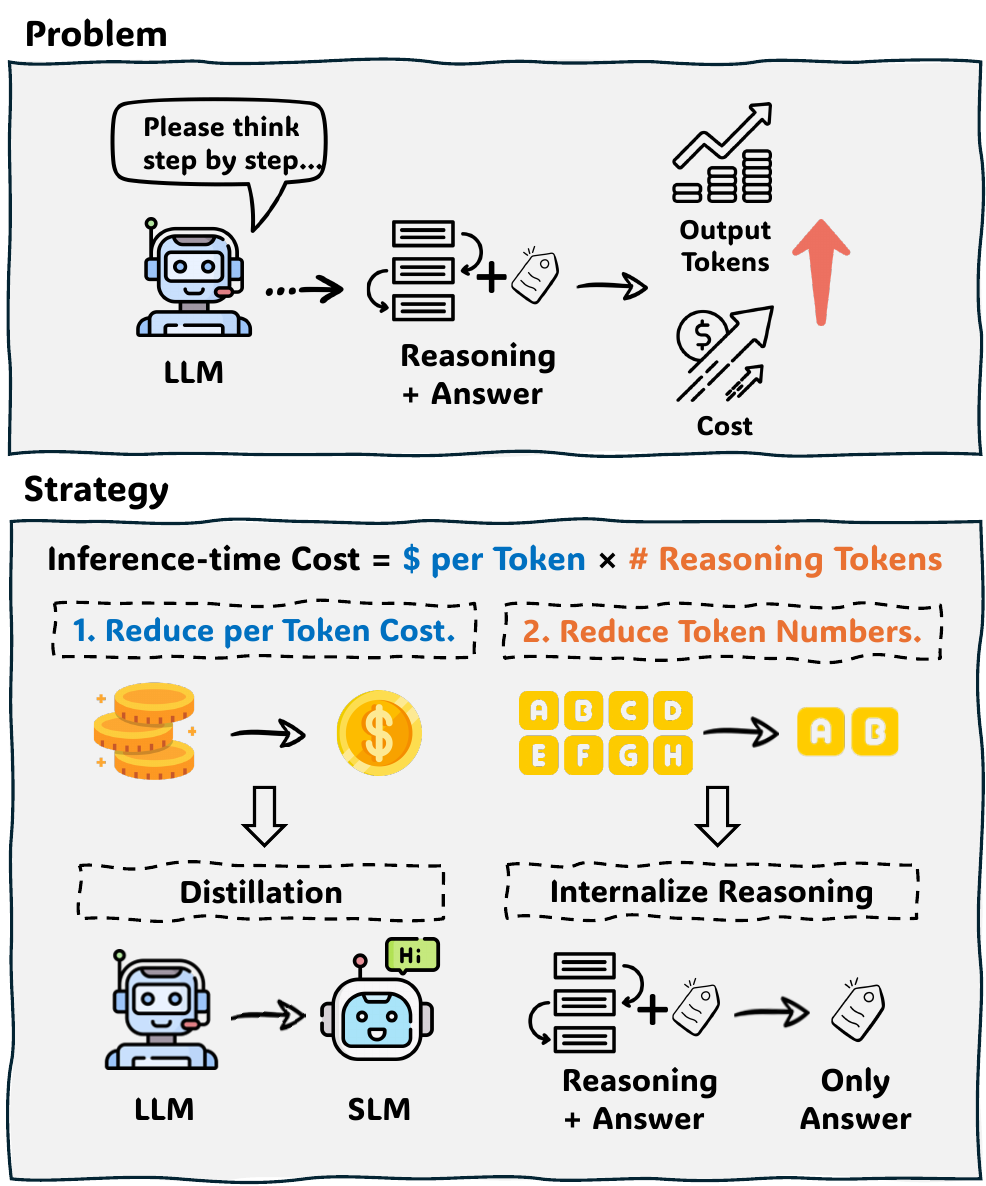} 
	\caption{\textbf{Overview of the Proposed Method.} The upper part of the figure illustrates the background problem: while generating more reasoning steps improves the performance of LLMs, it also leads to significantly higher computational costs. To mitigate this, we propose targeted strategies, shown in the lower part of the figure. Our approach reduces the cost per token by distilling knowledge from large models into smaller ones and minimizes the total number of tokens by gradually shrinking intermediate reasoning paths.} 
    \vspace{-4mm}
	\label{fig:intro-1} 
\end{figure}


Recent research has primarily explored two key approaches to addressing this issue: reducing the cost per token and reducing the number of reasoning tokens as illustrated in Figure \ref{fig:intro-1} \cite{wang2024reasoning,hsieh2023distilling,wang2024reducing}. A common strategy for reducing the cost per token is to replace large models with smaller, more efficient ones. However, directly using smaller models often results in a significant drop in performance, particularly on complex reasoning tasks. To mitigate this problem, knowledge distillation \cite{hinton2015distilling,park2019relational} has emerged as an effective solution, enabling student models to mimic the performance of larger teacher models \cite{tang2019distilling,hsieh2023distilling}. However, traditional knowledge distillation typically relies on expensive human-labeled data, limiting its practical applicability. Additionally, most distillation methods employ only a single teacher model, restricting the diversity of knowledge that could be leveraged from multiple teacher models. To overcome these limitations, we propose Dual-Criteria Rejection Sampling (DCRS), a method that first utilizes a multi-teacher strategy to generate pseudo-labels and then applies a two-stage selection process-Quality Selection and Diversity Selection-to construct a high-quality and diverse distillation dataset. This approach not only enhances the efficiency of knowledge transfer but also enables our model to adapt effectively to unsupervised settings.

For the other issue, the number of reasoning tokens can be reduced by shrinking intermediate reasoning paths \cite{wang2024reasoning,deng2024explicit}. While this effectively lowers token usage, it often comes at the cost of degraded model performance. Therefore, it is essential to develop a method that balances computational efficiency and model accuracy. Consider a real-world learning scenario where a teacher possesses a deep and comprehensive understanding of a concept, while a student may struggle to grasp the material fully. To bridge this gap, the teacher distills knowledge, extracting only the most essential and concise information to help the student learn more effectively. Over time, the student internalizes this reasoning process, enabling them to generate answers instantly upon encountering a question, without requiring explicit intermediate reasoning steps.

Inspired by this human cognitive process, we propose Habitual Reasoning Distillation (HaRD), a method that internalizes explicit reasoning into the model’s habitual behavior through a multi-stage distillation process, thereby reducing the need for explicit reasoning during inference. HaRD follows a three-stage distillation strategy: (a) Full Reasoning Distillation, where the student learns reasoning patterns from complete reasoning paths generated by teacher models; (b) Reasoning-Compressed Distillation, where the reasoning process is progressively compressed, with teachers refining their outputs based on the student’s responses to create reasoning paths aligned with the student’s capabilities; and (c) Reasoning-Free Distillation, where the student is trained without explicit reasoning steps, relying only on final labels, allowing it to generate high-quality answers directly. This process shifts the computational burden from inference to training, enabling both high performance and low inference cost.

In this work, we propose \textit{\textbf{TwT}} (Thinking without Tokens), a method that achieves an optimal balance between inference-time computational cost and performance. TwT follows a two-step process. First, DCRS utilizes multi-teacher LLMs to generate pseudo-labels, enabling the model to adapt to unsupervised settings. Then, HaRD applies a multi-stage distillation approach to progressively internalize explicit reasoning abilities into the student model as inherent capabilities. Our key contributions are summarized as follows:

\begin{itemize}
    \item \textbf{Novel Distillation Framework:} We propose \textit{\textbf{TwT}}, a novel framework that aims to reduce inference-time computational cost through habitual reasoning distillation with multi-teachers' guidance while preserving high performance.
    \item \textbf{Unsupervised Sampling Strategy:} We propose Dual-Criteria Rejection Sampling (DCRS), a method that selects high-quality and diverse distillation data generated by multi-teacher LLMs, enabling adaptation to unsupervised settings.
    \item \textbf{Efficient Reasoning Distillation:} We design a Habitual Reasoning Distillation (HaRD) method that refines reasoning patterns through a teacher-guided compression strategy, ensuring better alignment with the student model’s capabilities and ultimately integrates explicit reasoning into the model’s inherent behavior.
    \item \textbf{Comprehensive Empirical Validation:} Experimental results demonstrate that our approach outperforms existing distillation techniques, achieving up to a 13.6\% improvement in performance while generating fewer tokens.
\end{itemize}

\begin{figure*}[t] 
	\centering 
	\includegraphics[width=\textwidth]{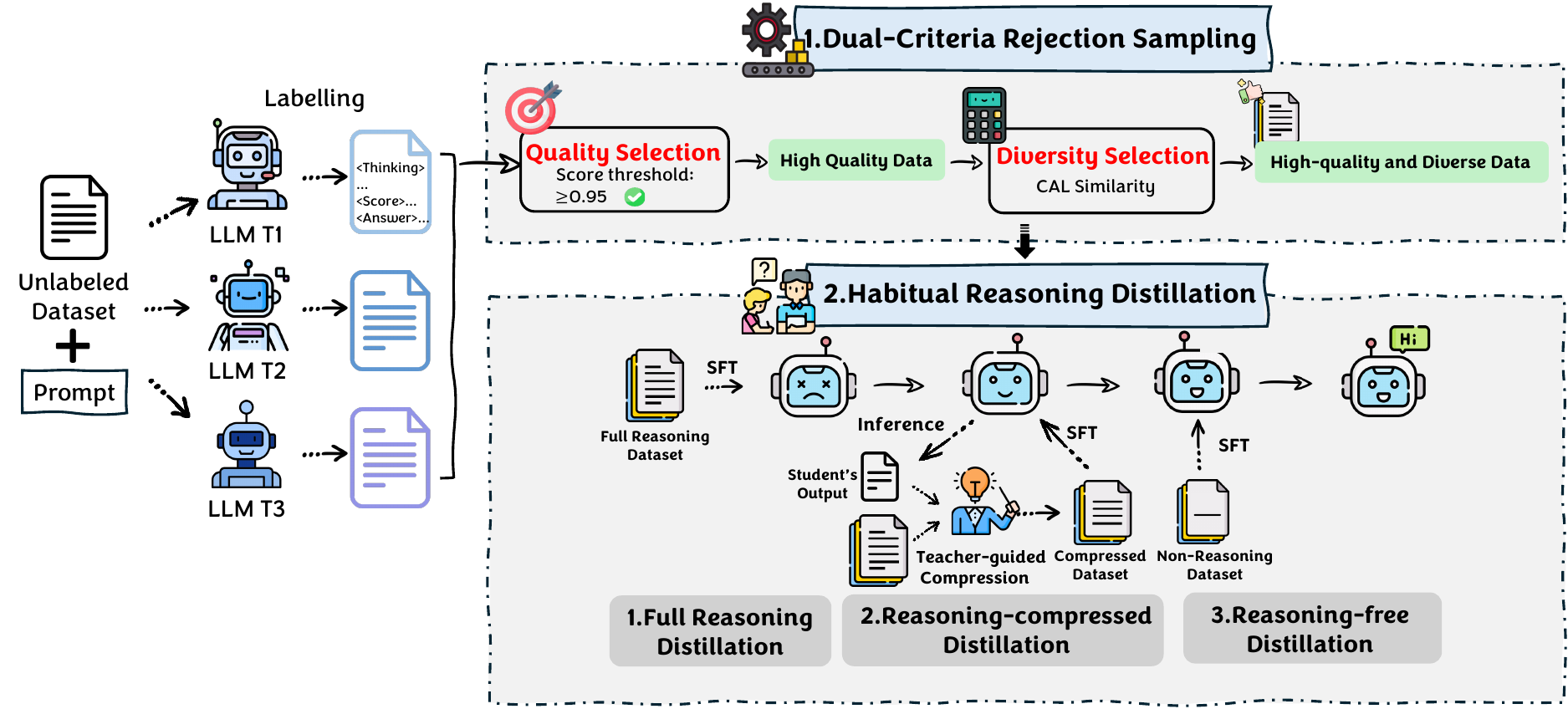} 
	\caption{\textbf{Method Framework.} Our proposed \textit{\textbf{TwT}} (Thinking without Tokens) framework consists of two stages: Dual-Criteria Rejection Sampling (\textbf{DCRS}) and Habitual Reasoning Distillation (\textbf{HaRD}). In the first stage, DCRS selects a high-quality and diverse reasoning distillation dataset generated by multiple teacher LLMs (e.g., T1, T2, T3). In the second stage, HaRD progressively internalizes reasoning ability into the student model through a three-stage distillation process.} 
	\label{workflow} 
    \vspace{-4mm}
\end{figure*}
\section{Related Work}

\subsection{Knowledge Distillation for LLMs}
Knowledge distillation (KD) \cite{hinton2015distilling,beyer2022knowledge,west2021symbolic} transfers capabilities from large LLMs to smaller ones using hard or soft labels from a teacher model. However, by relying solely on final outputs, standard KD provides limited information. Reasoning distillation \cite{hsieh2023distilling} addresses this by training students to understand both the final answer and the underlying reasoning. Recent work \cite{chen2023mcc,liu2023mind} explores generating multiple rationales per query to enhance robustness. However, using a single teacher can introduce bias and limit diversity. Multi-teacher strategies \cite{tian2024beyond,10697478} address this by aggregating diverse reasoning paths, enriching distillation data, and improving generalization. These methods rely on labeled datasets to select high-quality distilled data, which can be difficult to obtain.
Our work introduces a multi-teacher approach to incorporate diverse reasoning data and proposes a Dual-Criteria Rejection Sampling strategy to obtain a high-quality and diverse distillation dataset from unlabeled data.
\subsection{Reasoning and Inference-time Scaling}
Recent work enhances output diversity by improving reasoning via structural methods (e.g., code parsing, problem decomposition \cite{gao2023pal,zhouleast}) and by generating multiple reasoning paths through techniques like majority voting and reinforcement learning \cite{wei2022chain,yao2023tree,cao-2024-graphreason,wangself,fu2022complexity,huang2023large,trung2024reft,wang2024math}. However, these supervised approaches rely on a single model and labeled data, limiting inherent diversity. Additionally, studies such as \cite{snell2024scaling,wu2024inference} show that adaptive inference-time strategies can significantly reduce compute costs, but model performance will be reduced. To address these issues, we propose a multi-teacher strategy, an unsupervised approach for generating diverse, high-quality samples with habitual reasoning distillation for efficient inference with explicit reasoning.
\section{Method}
In this section, we provide a detailed explanation of the implementation of our TwT as illustrated in Fig. \ref{workflow}. First, we propose a Dual-Criteria Rejection Sampling (DCRS) strategy to obtain high-quality and diverse distillation samples (Section 3.1). Then, we design a \textbf{Ha}bitual \textbf{R}easoning \textbf{D}istillation (HaRD) strategy that progressively internalizes the reasoning ability at each distillation stage, allowing the reasoning capabilities of the teacher models to be gradually internalized into the student model (Section 3.2).
\begin{figure*}[t] 
	\centering 
	\includegraphics[width=\textwidth]{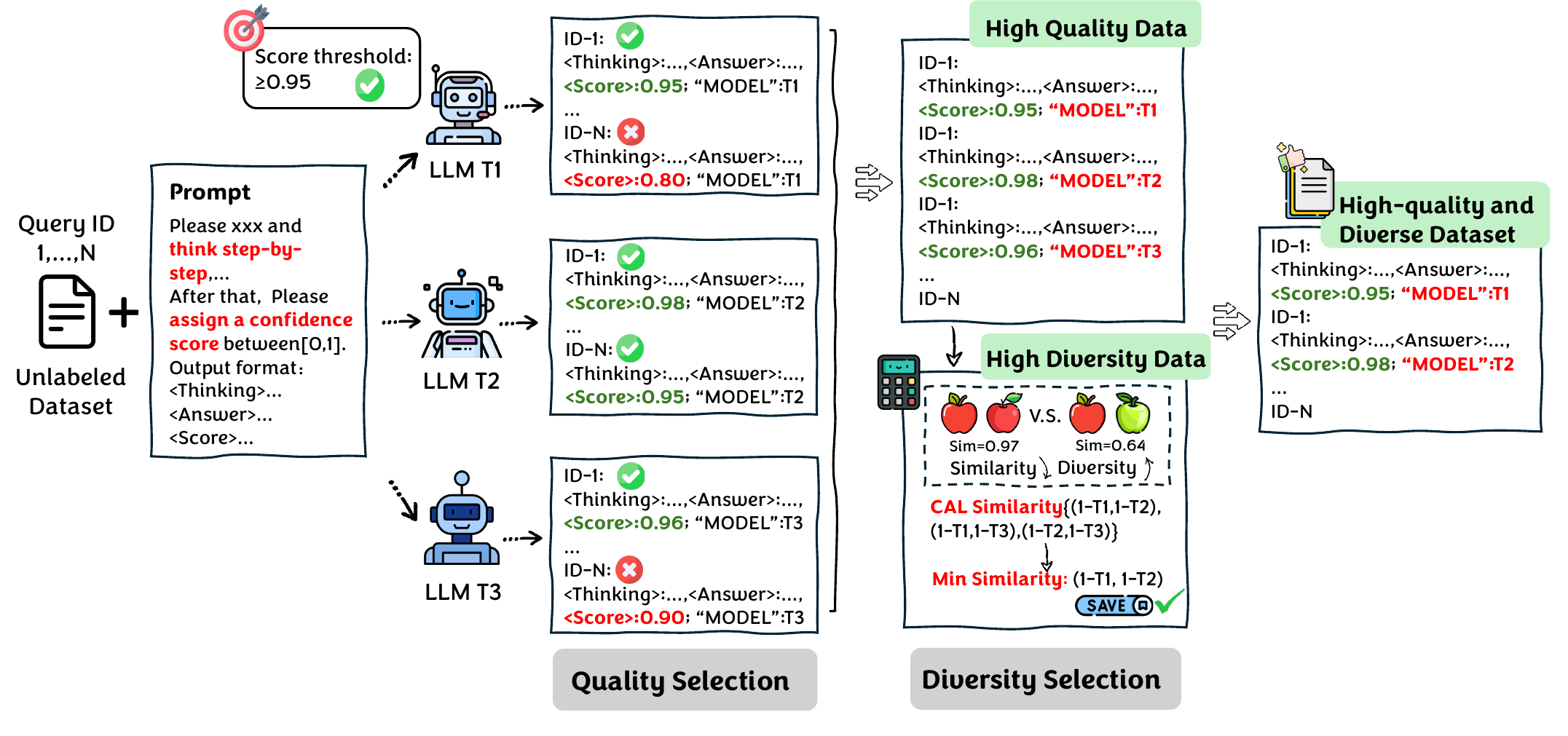} 
	\caption{\textbf{Dual-Criteria Rejection Sampling Architecture.} Our proposed DCRS method comprises two stages: Quality Selection and Diversity Selection. The first stage filters samples using confidence scores, while the second stage enhances diversity by selecting samples based on similarity. This approach ensures a high-quality and diverse distillation dataset, enabling our method to effectively adapt to unsupervised scenarios.} 
	\label{Fig.feature} 
    \vspace{-4mm}
\end{figure*}
\subsection{Dual-Criteria Rejection Sampling}
To provide the student model with high-quality and diverse reasoning paths, we propose a novel paradigm termed Dual-Criteria Rejection Sampling (DCRS), which extends traditional rejection sampling \cite{gilks1992adaptive} by integrating two key selection metrics: confidence scores and similarity measures. Leveraging a multi-teacher strategy, we first prompt teacher LLMs to generate an initial pool of pseudo-labels. DCRS performs sample selection in two sequential steps: quality selection and diversity selection. 



\subsubsection{Quality Selection}

As shown in Figure \ref{Fig.feature}, we take use of CoT prompting \cite{wei2022chain} to generate and extract reasoning patterns from multi-teacher LLMs. Given an unlabeled dataset $\mathcal{D} = \{ x_i \}_{i=1}^N$, where each $x_i$ is a query, we first design a prompt template $p$ to clarify the task solution method. The prompt instructs the LLM to produce the output $O_i$ in the form of a triplet $(r_i, y_i, c_i)$, where $y_i$ is the predicted label for task $x_i$, $r_i$ is the rationale provided by the LLM, and $c_i$ is the confidence score for the reasoning and predicted label, represented as a decimal in the range $[0, 1]$. To ensure that the confidence score is more reliable, we compute it using a weighted combination of multiple performance factors:
\begin{equation}
    c_i = \sum_{j=1}^n w_j \cdot m_j
\end{equation}
where $w_j$ denotes the weight assigned to the $j$-th factor and $m_j$ represents the corresponding factor value. For example, in a code generation task, we let the large model assign a weighted score based on factors such as the reasoning process, code readability, and robustness, yielding the final confidence score. For $ k$ teacher models, we can obtain a set of outputs $O_i = \{ O_{i1}, O_{i2}, \dots, O_{ik} \}$ for each query $x_i$. Then, we set a confidence score threshold $s$. For each output $O_i$, if its confidence score $c_i\ge s$, the output sample is considered high-quality and retained; otherwise, it is discarded. Subsequently, diversity selection is performed on the retained sample set $\mathcal{H} = \{ (x_i, r_i, y_i,c_i) \}_{i=1}^{P}$, where $P$ is the total number of retained samples.

\subsubsection{Diversity Selection}
For the high-quality sample set $\mathcal{H}$, if the outputs $O_i = \{O_{i1}, O_{i2}, \dots, O_{ik} \}$ for the same query $x_i$ come from three or more different teacher models (i.e., $k \geq 3$), we calculate the semantic similarity between the rationales provided by these teacher models. Specifically, we assume that $r_{ij}$ denotes the rationale generated by the $j$-th teacher for query $x_i$. Each rationale $r_{ij}$ is then mapped into a fixed-dimensional embedding $E(r_{ij})$ using a pre-trained sentence embedding model. The cosine similarity between any two rationales $r_{ip}$ and $r_{iq}$ is computed as:
\begin{equation}
\text{sim}(r_{ip}, r_{iq}) = \frac{E(r_{ip}) \cdot E(r_{iq})}{\|E(r_{ip})\| \, \|E(r_{iq})\|}
\end{equation}
where $r_{ip}$ and $r_{iq}$ are rationales provided by two distinct teacher models for the same question, with $1 \leq p < q \leq k$. We calculate the cosine similarity for all unique pairs $(r_{ip}, r_{iq})$. To maximize diversity, we select the pair of rationales that yields the lowest similarity score. In our implementation, we directly select the pair of rationales that exhibits the minimum cosine similarity. For a given query $x_i$, we define:
\begin{equation}
(p^*, q^*) = \arg\min_{1 \leq p < q \leq k} \text{sim}\bigl(r_{ip}, r_{iq}\bigr)  
\end{equation}
 where $(p^*, q^*)$ are the indices corresponding to the pair of rationales with the smallest similarity, thereby ensuring that the final distillation dataset $\mathcal{G} = \{ (x_i, r_i, y_i) \}_{i=1}^{Q}$ is both high-quality and diverse, where $P$ is the total number of final distillation samples.

\subsection{Habitual Reasoning Distillation}


In this section, we introduces a novel multi-stage distillation strategy designed to balance inference-time efficiency and model performance. The process is structured into three sequential stages. In Stage-1, the student model is trained on data with full reasoning, enabling it to internalize the teacher’s comprehensive problem-solving process. In Stage-2, the student learns from compressed reasoning, fostering more concise and efficient inference. In Stage-3, distillation is performed using answer-only data, encouraging the student to establish a direct query-to-answer mapping.
Crucially, Stage-2 incorporates a Teacher-Guided Compression mechanism that adaptively aligns the reasoning complexity with the student’s capacity.

\paragraph{Stage-1: Full Reasoning Distillation.} In this stage, the student model is trained to learn the complete reasoning paths under the supervision of the teacher models' full reasoning. The goal is to help the weak student model understand the logical steps involved in the task and build a solid foundation for further distillation.

\definecolor{langgreen}{rgb}{0.0, 0.6, 0.3}

\begin{table*}[htb]
\footnotesize
\centering
\setlength{\tabcolsep}{1.6mm}{
\begin{tabularx}{\textwidth}{cc|cccccc}
\toprule
\multicolumn{2}{c|}{\multirow{2}*{\textbf{Method}}}
& \multicolumn{2}{c}{\textbf{MBPP}} 
& \multicolumn{2}{c}{\textbf{CQA}} 
& \multicolumn{2}{c}{\textbf{MetaMath}} \\
\cmidrule(r){3-4} \cmidrule(r){5-6} \cmidrule(r){7-8}
& & \textbf{Pass@1} & \textbf{Token} 
& \textbf{Accuracy} & \textbf{Token} 
& \textbf{Accuracy} & \textbf{Token} \\
\midrule
\multicolumn{2}{c|}{GPT-4} 
& 77.68\% &493 &83.27\% &312 &86.31\% &512  \\
\multicolumn{2}{c|}{GPT-4o-mini} 
& 77.14\% &451 &81.76\% &301 &87.03\% &529 \\
\multicolumn{2}{c|}{Mistral-large} 
& 73.83\% &365 &80.15\% &288 &86.33\% &463 \\
\midrule
\multirow{5}{*}{Mistral-7B-v0.3} 
& Vanilla Student & 42.90\% & 209 & 60.19\% & 171 & 30.12\% & 255 \\
& Standard KD    & 52.67\% & \underline{53} & 63.63\% & \textbf{5} & 38.19\% & \textbf{7} \\
& Distilling     & 53.39\% & 170 & 64.47\% & 129 & \underline{42.20\%} & 397 \\
& Tinyllm        & \underline{54.45\%} & 175 & \underline{67.39\%} & 225 & 41.67\% & \underline{226} \\
\rowcolor{gray!20}
& \textbf{TwT} 
  & \textbf{57.11\%} \textcolor{langgreen}{(\scalebox{0.8}{$\uparrow4.88\%$})} 
  & \textbf{48}  
  & \textbf{76.16\%} \textcolor{langgreen}{(\scalebox{0.8}{$\uparrow13.01\%$})} 
  & \underline{6} 
  & \textbf{47.94\%} \textcolor{langgreen}{(\scalebox{0.8}{$\uparrow13.60\%$})} 
  & \textbf{7} \\
\midrule
\multirow{5}{*}{Phi-3.5mini} 
& Vanilla Student & 54.71\% & 199 & 64.15\% & 150 & 73.36\% & 279 \\
& Standard KD    & 61.71\% & \textbf{98} & 64.48\% & \textbf{6} & 78.16\% & \textbf{10} \\
& Distilling     & 62.03\% & 202 & 65.74\% & 148 & 78.19\% & 370 \\
& Tinyllm        & \underline{64.44\%} & 192 & \underline{70.30\%} & 144 & \underline{78.23\%} & 240 \\
\rowcolor{gray!20}
& \textbf{TwT} 
  & \textbf{67.93\%} \textcolor{langgreen}{(\scalebox{0.8}{$\uparrow5.42\%$})} 
  & \underline{105} 
  & \textbf{76.42\%} \textcolor{langgreen}{(\scalebox{0.8}{$\uparrow8.70\%$})} 
  & \underline{10} 
  & \textbf{83.58\%} \textcolor{langgreen}{(\scalebox{0.8}{$\uparrow6.84\%$})} 
  & \underline{15} \\
\bottomrule
\end{tabularx}
}
\caption{\textbf{Quantitative results for baseline models.} The top three rows show the inference results of our teacher models, while “Distilling” is an abbreviation for “Distilling Step-by-Step.” The best and the second best results are highlighted in bold and underlined respectively. The improvements of TwT over the second best results are shown in green with an upward arrow.}
\vspace{-4mm}
\label{tab:mainresults}
\end{table*}


The teachers' reasoning ability can be transferred by fine-tuning the student model using the full reasoning $\mathcal{G}$ derived from high-quality and diversity sample selection. More specifically, the process of learning full reasoning paths through fine-tuning is defined as follows:
\begin{equation}
\mathcal{L}_{\mathrm{1}}=\mathbb{E}_\mathcal{G} [ \log \mathrm{P}_{f}([x ; r ; y])]
\end{equation}
where $f$ indicates the student model, $\mathbb{E}_\mathcal{G}$ is the expectation over the distillation dataset $\mathcal{G}$, and $\mathrm{P}_{f}([x ;  r ; \hat{y}])$ is  the probability assigned by the student model f to the joint input $[x;  r;  \hat{y}]$.

\paragraph{Stage-2: Reasoning-Compressed Distillation.} In this stage, we progressively simplify the reasoning paths of the teacher model, generating more concise one by compressing the original reasoning paths. 
We observed that for the same problem, the outputs of student models are often shorter and feature more concise reasoning steps compared to those of teacher models. Therefore, we adopt a Teacher-Guided Compression approach that ensures the reasoning paths provided by the teacher models are better aligned with the learning characteristics of the student model, thereby enhancing overall distillation performance. 

Specifically, for a given query $x_i$, the teacher model generates the original reasoning $r_i$, while the student model produces the reasoning $r^{s1}_i$. We design a prompt $p'$ to guide the teacher model in refining its original reasoning $r_i$ based on the characteristics of the student model’s output (e.g., output length, complexity of understanding the problem). This process can be represented as $(p', r_i, r^{s1}_i)$ $\rightarrow r^{T}_i$. Subsequently, we replace the original reasoning $r_i$ in the dataset $\mathcal{G} = \{ (x_i, r_i, y_i) \}_{i=1}^{N}$ with the refined reasoning $r^{T}_i$, resulting in the second-stage distillation dataset $\mathcal{G'} = \{ (x_i, r^{T}_i, y_i) \}_{i=1}^{N}$. The second stage fine-tuning can be defined as:
\begin{equation}
\mathcal{L}_{\mathrm{2}}=\mathbb{E}_{\mathcal{G'}}[ \log \mathrm{P}_{f}([x ; r^T ; y])]
\end{equation}

\begin{table*}[]
\footnotesize
\setlength{\tabcolsep}{0.8mm}{
\begin{tabularx}{\textwidth}{c|cccccccccccc}
\toprule
\multirow{3}{*}{\textbf{Method}}& \multicolumn{4}{c}{\textbf{MBPP}}                                                                                             & \multicolumn{4}{c}{\textbf{CQA}}                                                        & \multicolumn{4}{c}{\textbf{MetaMath}}                                                   \\ 
\cmidrule(l){2-5} \cmidrule(l){6-9} \cmidrule(l){10-13} 
& \multicolumn{2}{c}{\textbf{Mistral-7B-v0.3}}                  & \multicolumn{2}{c}{\textbf{Phi-3.5mini}}                      & \multicolumn{2}{c}{\textbf{Mistral-7B-v0.3}} & \multicolumn{2}{c}{\textbf{Phi-3.5mini}} & \multicolumn{2}{c}{\textbf{Mistral-7B-v0.3}} & \multicolumn{2}{c}{\textbf{Phi-3.5mini}} \\ 
\cmidrule(l){2-3} \cmidrule(l){4-5} \cmidrule(l){6-7} \cmidrule(l){8-9} \cmidrule(l){10-11} \cmidrule(l){12-13} 
 & {\textbf{Pass@1}} & \textbf{Token} & {\textbf{Pass@1}} & \textbf{Token} & \textbf{Accuracy}      & \textbf{Token}      & \textbf{Accuracy}    & \textbf{Token}    & \textbf{Accuracy}      & \textbf{Token}      & \textbf{Accuracy}    & \textbf{Token}    \\ \midrule
TwT-stage1                        & 54.83\%                                      & 291            & 65.32\%                                      & 310            & 73.31\%                & 141                 & 72.99\%              & 138               & 46.48\%                & 295                 & 80.05\%              & 313               \\
TwT-stage2                        & 56.48\%                                      & 154            & 66.42\%                                      & 184            & 74.89\%                & 84                  & 75.39\%              & 73                & 47.02\%                & 169                 & 81.33\%              & 196               \\
\rowcolor{gray!20} 
\textbf{TwT-stage3}               & \textbf{57.11\%}                             & \textbf{48}    & \textbf{67.93\%}                             & \textbf{105}   & \textbf{76.16\%}       & \textbf{8}          & \textbf{76.42\%}     & \textbf{10}       & \textbf{47.94\%}       & \textbf{12}         & \textbf{83.58\%}     & \textbf{15}       \\ \bottomrule
\end{tabularx}
}
\caption{\textbf{Quantitative results for three distillation stages.} Accuracy and the number of output tokens were used to evaluate the model performance. The best results were highlighted in bold.}
\label{tab:distillation3}
\end{table*}

\paragraph{Stage-3: Reasoning-Free Distillation.} Finally, we completely remove the reasoning paths and only retain the final answer as the supervision signal. The student model is trained to directly output the correct answer without relying on any reasoning chain. The goal of this stage is to enable the student model to form a “habitual” ability, allowing it to efficiently complete tasks without the need for complex reasoning. After removing $r$, the new dataset $\mathcal{G''} = \{ (x_i, y_i) \}_{i=1}^{N}$ only contains the original query $x_i$ and the label $y_i$ predicted by the teacher model. The third stage fine-tuning can be defined as:
\begin{equation}
\mathcal{L}_{\mathrm{3}}=\mathbb{E}_{\mathcal{G''}} [\log \mathrm{P}_{f}([x ;  y])]
\end{equation}

In summary, Stage-1 uses full reasoning distillation, while Stage-2 employs compressed reasoning distillation, enabling the student model to learn complete reasoning chains and establish systematic reasoning patterns. Stage-3 conducts end-to-end training to strengthen the model’s understanding of problem-answer relationships, with explicit reasoning now internalized. This progressive approach enhances performance while reducing inference-time computation. The detailed prompt and case study are provided in Appendix \ref{appendix:prompt}
and \ref{appendix:case}.

\section{Experiment}

\begin{table*}[]
\footnotesize
\setlength{\tabcolsep}{0.8mm}{
\begin{tabularx}{\textwidth}{c|cccccccccccc}
\toprule
\multirow{3}{*}{\textbf{Method}}& \multicolumn{4}{c}{\textbf{MBPP}}                                                                                             & \multicolumn{4}{c}{\textbf{CQA}}                                                        & \multicolumn{4}{c}{\textbf{MetaMath}}                                                   \\ 
\cmidrule(l){2-5} \cmidrule(l){6-9} \cmidrule(l){10-13}
& \multicolumn{2}{c}{\textbf{Mistral-7B-v0.3}}                  & \multicolumn{2}{c}{\textbf{Phi-3.5mini}}                      & \multicolumn{2}{c}{\textbf{Mistral-7B-v0.3}} & \multicolumn{2}{c}{\textbf{Phi-3.5mini}} & \multicolumn{2}{c}{\textbf{Mistral-7B-v0.3}} & \multicolumn{2}{c}{\textbf{Phi-3.5mini}} \\ 
\cmidrule(l){2-3} \cmidrule(l){4-5} \cmidrule(l){6-7} \cmidrule(l){8-9} \cmidrule(l){10-11} \cmidrule(l){12-13} 
 & {\textbf{Pass@1}} & \textbf{Token} & {\textbf{Pass@1}} & \textbf{Token} & \textbf{Accuracy}      & \textbf{Token}      & \textbf{Accuracy}    & \textbf{Token}    & \textbf{Accuracy}      & \textbf{Token}      & \textbf{Accuracy}    & \textbf{Token}    \\ \midrule
TwT-stage1                       & 54.83\%                                      & 291            & 65.32\%                                      & 310            & 73.31\%                & 141                 & 72.99\%              & 138               & 46.48\%                & 295                 & 80.05\%              & 313               \\
TwT-stage2                       & 56.48\%                                      & 154            & 66.42\%                                      & 184            & 74.89\%                & 84                  & 75.39\%              & 73                & 47.02\%                & 169                 & 81.33\%              & 196               \\

{TwT-stage3}              & {56.94\%}                             & {133}   & {67.18\%}                             & {161}   & {75.44\%}       & {55}         & {75.85\%}     & {64}       & {47.41\%}       & {144}        & {81.99\%}     & {172}  \\ 
\rowcolor{gray!20} 
\textbf{TwT-stage4}                                  & \textbf{57.49\%}                             & \textbf{42}    & \textbf{68.21\%}                             & \textbf{100}   & \textbf{76.29\%}       & \textbf{8}          & \textbf{76.88\%}     & \textbf{9}        & \textbf{48.37\%}       & \textbf{12}         & \textbf{83.61\%}     & \textbf{13}      \\

\bottomrule   
\end{tabularx}
}
\caption{\textbf{Quantitative results for four distillation stages.} Accuracy and the number of output tokens were used to evaluate the model performance. The best results were highlighted in bold.}
\label{tab:distillation4}
\end{table*}

\begin{table*}[]
\footnotesize
\setlength{\tabcolsep}{0.8mm}{
\begin{tabularx}{\textwidth}{c|cccccccccccc}
\toprule
\multirow{3}{*}{\textbf{Method}}& \multicolumn{4}{c}{\textbf{MBPP}}                                                                                             & \multicolumn{4}{c}{\textbf{CQA}}                                                        & \multicolumn{4}{c}{\textbf{MetaMath}}                                                   \\ 
\cmidrule(l){2-5} \cmidrule(l){6-9} \cmidrule(l){10-13} 
& \multicolumn{2}{c}{\textbf{Mistral-7B-v0.3}}                  & \multicolumn{2}{c}{\textbf{Phi-3.5mini}}                      & \multicolumn{2}{c}{\textbf{Mistral-7B-v0.3}} & \multicolumn{2}{c}{\textbf{Phi-3.5mini}} & \multicolumn{2}{c}{\textbf{Mistral-7B-v0.3}} & \multicolumn{2}{c}{\textbf{Phi-3.5mini}} \\ 
\cmidrule(l){2-3} \cmidrule(l){4-5} \cmidrule(l){6-7} \cmidrule(l){8-9} \cmidrule(l){10-11} \cmidrule(l){12-13} 
 & {\textbf{Pass@1}} & \textbf{Token} & {\textbf{Pass@1}} & \textbf{Token} & \textbf{Accuracy}      & \textbf{Token}      & \textbf{Accuracy}    & \textbf{Token}    & \textbf{Accuracy}      & \textbf{Token}      & \textbf{Accuracy}    & \textbf{Token}    \\ \midrule
TwT-stage1                        & 54.56\% & 306 & 64.10\% & 297 & 72.91\% & 155 & 72.49\% & 141 & 45.09\% & 299 & 79.17\% & 320 \\
TwT-stage2                        & 55.31\% & 162 & 65.93\% & 188 & 75.22\% & 72  & 75.62\% & 74  & 46.29\% & 176 & 80.62\% & 187 \\
\rowcolor{gray!20} 
\textbf{TwT-stage3}               & \textbf{56.64\%} & \textbf{57} & \textbf{66.16\%} & \textbf{113} & \textbf{75.98\%} & \textbf{12} & \textbf{76.15\%} & \textbf{15} & \textbf{46.93\%} & \textbf{16} & \textbf{81.37\%} & \textbf{18} \\ 
\bottomrule
\end{tabularx}
}
\caption{\textbf{Quantitative results for three distillation stages via GPT-4, GPT-4o-mini and GPT-3.5-turbo.} Accuracy and the number of output tokens were used to evaluate the model performance. The best results were highlighted in bold.}
\label{tab:distillation_results}
\end{table*}

\subsection{Experiment Setup}
\noindent \textbf{Datasets.} We evaluate our TwT on 3 benchmark datasets for 3 different NLP tasks: MBPP \cite{austin2021program} for NL to python code generalization; CommonsenseQA (CQA) \cite{talmor2018commonsenseqa} for commonsense question answering; MetaMathQA (MetaMath) \cite{yu2023metamath} for mathematical reasoning, which is augmented from the training sets of GSM8K \cite{cobbe2021training} and MATH \cite{hendrycks2021measuring}.

\noindent \textbf{Models.} We utilize GPT-4, GPT-4o-mini and Mistral-Large as our teacher models, which are accessed through OpenAI’s API and MistralAI's API. For the student models, we use Mistral-7B-v0.3 and Phi-3.5mini. For the pre-trained sentence embedding model, we leverage all-mpnet-base-v2.

\noindent \textbf{Baselines.} For our baselines, we evaluate three types of methods: teacher model's performance, vanilla student model's performance, and knowledge distillation based methods that containing Standard-KD \cite{hinton2015distilling}, a general distillation method that fine-tunes the student model using the teacher model’s generated labels as ground-truth; Distilling-Step-by-Step \cite{hsieh2023distilling}, which leverages LLM-generated rationales as additional supervision to train smaller models; TinyLLM \cite{tian2024beyond}, a paradigm that distills diverse reasoning paths from multiple teacher LLMs into a student model.

\noindent \textbf{Implementation Details.} In our experimental setup, we employed training with LoRA fine-tuning, setting the LoRA rank to 8, a learning rate of 1e-5, a batch size of 8, 4 training epochs, and a context window of 4096 tokens. During inference, we used a temperature of 0, max tokens set to 2048, and a top-p value of 0.95. For the sampling process, we selected a scoring threshold of $s=0.95$. All experiments were conducted on four NVIDIA A100 Tensor Core GPUs, enabling large-scale training and efficient computation. The prompts for specific methods and a case study are provided in the appendix.

\subsection{Baseline Comparison}

Across three specific-tasks, TwT consistently outperforms other distillation methods as shown in Table \ref{tab:mainresults}. Compared with the best performing baseline, TwT achieves an improvement of up to 13.60\% compared to the best-performing baseline, while reducing the token number by 98.2\% (token numbers from 397 to 7 on MetaMath dataset), substantially lowering the inference cost. Typically, reducing inference tokens leads to a decline in performance; however, TwT overcomes this trade off by maintaining or even enhancing model performance while dramatically reducing token usage, thus achieving both high performance and low inference-time computational cost simultaneously. In addition, TwT effectively bridges the gap between the student and teacher models, significantly narrowing the performance disparity observed in vanilla student models.

\subsection{Analysis on Distillation Stage}
In the Habitual Reasoning Distillation phase, we separately evaluated the student model’s performance at each fine-tuning stage and tracked the number of output tokens during inference, as shown in Table \ref{tab:distillation3}. The results indicate that TwT steadily improves with each stage, while the inference token numbers gradually decrease. By leveraging our distillation strategy, the model successfully internalizes the reasoning process as part of its own capabilities.   

Furthermore, we expanded our three-stage procedure by extending it to four stages, adding an additional step after the second stage where the teacher model further compresses the reasoning process based on the student’s output. As shown in Table \ref{tab:distillation4}, this extension yields a slight improvement but does not significantly surpass the three-stage method. This suggests that our original three-stage process is already effective in achieving the desired performance.

\begin{table}[]
\footnotesize
\centering
\setlength{\tabcolsep}{0.5mm}{
\begin{tabularx}{0.45\textwidth}{c|ccc}
\toprule
\textbf{Samping Method}                & \textbf{MBPP}    & \textbf{CQA}     & \textbf{MetaMath} \\ \midrule
Log Probability         & 74.83\% & 75.29\% & 72.19\%  \\
Hard Rejection Sampling & 72.29\% & 73.11\% & 70.26\%  \\
\rowcolor{gray!20}
\textbf{Confidence Score}        & \textbf{83.49\%} & \textbf{84.90\%} & \textbf{81.14\%}  \\ \bottomrule
\end{tabularx}
}
\caption{\textbf{Quantitative results for sampling methods.} Accuracy was used to evaluate the model performance. The best results were highlighted in bold.}
\label{tab:sampling}
\end{table}

\begin{table*}[]
\footnotesize
\setlength{\tabcolsep}{0.6mm}{
\begin{tabularx}{\textwidth}{c|cccccc}
\toprule
\multirow{2}{*}{\textbf{Compression Method}} & \multicolumn{2}{c}{\textbf{MBPP} }           & \multicolumn{2}{c}{\textbf{CQA} }            & \multicolumn{2}{c}{\textbf{MetaMath}}        \\ 
\cmidrule(l){2-3} \cmidrule(l){4-5} \cmidrule(l){6-7} 
& \textbf{Mistral-7B-v0.3}  & \textbf{Phi-3.5mini }     & \textbf{Mistral-7B-v0.3}  & \textbf{Phi-3.5mini}      & \textbf{Mistral-7B-v0.3}  & \textbf{Phi-3.5mini }     \\ \midrule
Fixed-Length Compression            & 55.89\%          & 64.94\%          & 73.29\%          & 72.92\%          & 43.18\%          & 80.62\%          \\
Compressor                          & 56.01\%          & 65.10\%          & 73.61\%          & 73.69\%          & 45.22\%          & 80.95\%          \\
\rowcolor{gray!20} \textbf{Teacher-guided Compression} & \textbf{56.48\%} & \textbf{66.42\%} & \textbf{74.89\%} & \textbf{75.39\%} & \textbf{47.02\%} & \textbf{81.33\%} \\
\bottomrule
\end{tabularx}
}
\caption{\textbf{Quantitative Results for Compression Methods.} Accuracy was used to evaluate the model performance. The best results were highlighted in bold.}
\label{tab:compression}
\end{table*}

\begin{table*}[]
\footnotesize
\setlength{\tabcolsep}{3.6mm}{
\begin{tabularx}{\textwidth}{c|cccccc}
\toprule
\multirow{2}{*}{\textbf{Methods}}  & \multicolumn{2}{c}{\textbf{MBPP}}  & \multicolumn{2}{c}{\textbf{CQA}}   & \multicolumn{2}{c}{\textbf{MetaMath}} \\ 
\cmidrule(l){2-3} \cmidrule(l){4-5} \cmidrule(l){6-7} 
& \textbf{Accuracy} & \textbf{Token} & \textbf{Accuracy} & \textbf{Token} & \textbf{Accuracy}   & \textbf{Token}  \\ \midrule
w/o Multi-Teacher Strategy (GPT-4) & 55.38\%           & 54             & 74.64\%           & 7              & 46.82\%             & 15              \\
w/o DCRS                           & 55.42\%           & 48             & 74.79\%           & 9              & 47.11\%             & 14              \\
w/o Compression Distillation Stage & 54.49\%           & 93             & 72.19\%           & 21             & 46.38\%             & 48              \\
\rowcolor{gray!20}
\textbf{TwT}                       & \textbf{57.11\%}  & \textbf{48}    & \textbf{76.16\%}  & \textbf{8}     & \textbf{47.94\%}    & \textbf{12}     \\ \bottomrule
\end{tabularx}
}
\caption{\textbf{Ablation Study for Model Components.} Accuracy and the number of output tokens were used to evaluate the model performance. The best results were highlighted in bold.}
\label{tab:component}
\end{table*}

\subsection{Analysis on the Robustness to Teacher Model Quality}

To evaluate the robustness of our method to the quality of teacher models, we conduct the experiment using teachers with varying capabilities. Specifically, we compare TwT performance when distilled from GPT-4, GPT-4o-mini, and GPT-3.5-turbo. Notably, GPT-3.5-turbo underperforms compared to Mistral-large on our benchmark datasets, particularly on the MetaMath task. Compared Table \ref{tab:distillation3} and Table \ref{tab:distillation_results}, it demonstrates that TwT remains effective even when guided by a weaker teacher. Despite the reduced capability of GPT-3.5-turbo, the resulting student models still follow our core hypothesis and achieve competitive performance. These findings suggest that TwT is robust to teacher quality and does not rely on the strongest teacher models to be effective.

\subsection{Analysis on Sampling and Compression Strategy}
We evaluated the effectiveness of our DCRS method’s confidence score selection for distillation data and our Teacher-Guided Compression approach for enhancing student model's performance. For sampling, Table \ref{tab:sampling} shows that our Confidence Score-Based method outperformed Log Probability-Based and Hard Rejection Sampling approaches by nearly 10\% in accuracy, demonstrating superior quality selection for distillation. For compression, Table \ref{tab:compression} shows that Teacher-Guided Compression better matched teacher outputs to student capacity than Fixed-Length or Compressor-based methods, improving both efficiency and performance. Confidence score details are in Appendix \ref{appendix:confidence}.

\subsection{Ablation Study}

We further analyze the contribution of each component to TwT’s performance through ablation studies. Specifically, w/o Multi-Teacher Strategy evaluates the effect of using a single teacher model, w/o DCRS evaluates performance without filtering the distillation data, and w/o Compression Distillation Stage analyzes the impact of directly removing the reasoning step as shown in Table \ref{tab:component}.

TwT's multi-teacher strategy outperforms single-teacher distillation by 1.4\%, demonstrating the benefit of diverse reasoning paths. Compared to using raw pseudo-labeled data directly, TwT's DCRS strategy shows a 1.5\% improvement, highlighting the value of quality data sampling. Additionally, TwT's multi-stage distillation achieves 3.5\% higher accuracy than uncompressed reasoning while reducing output tokens, proving the effectiveness of gradual reasoning internalization.

\section{Conclusion and Future Work}
We introduced TwT, a novel distillation framework that internalizes reasoning abilities into a student model under multi-teachers' guidance. It incorporates a Dual-Criteria Rejection Sampling stage to obtain a high-quality and diverse distillation dataset and a Habitual Reasoning Distillation strategy to gradually integrate reasoning capabilities into the student model. TwT achieves high performance with low inference cost without relying on labeled data or an explicit reasoning process.  In future work, we will continue to explore whether further subdividing the distillation stages can enhance our framework

\section*{Limitations}
Although our method has achieved excellent results, there are still some minor flaws here. One  limitation of our approach is that it currently only works effectively on specific tasks and is not applicable to datasets containing mixed tasks. Additionally, the Dual-Criteria Rejection Sampling process could consist of noise. The impact of this potential noise on performance is still undetermined. A potential future direction is to investigate implicit natural language reasoning by utilizing more advanced training strategies. While current tasks are primarily focused on explicit reasoning, incorporating implicit reasoning mechanisms could improve the model's robustness and its ability to generalize across different tasks.

\bibliography{custom}

\clearpage
\appendix


\section{Prompt}
\label{appendix:prompt}

\begin{figure*}[!htb] 
	\centering 
	\includegraphics[width=0.76\textwidth]{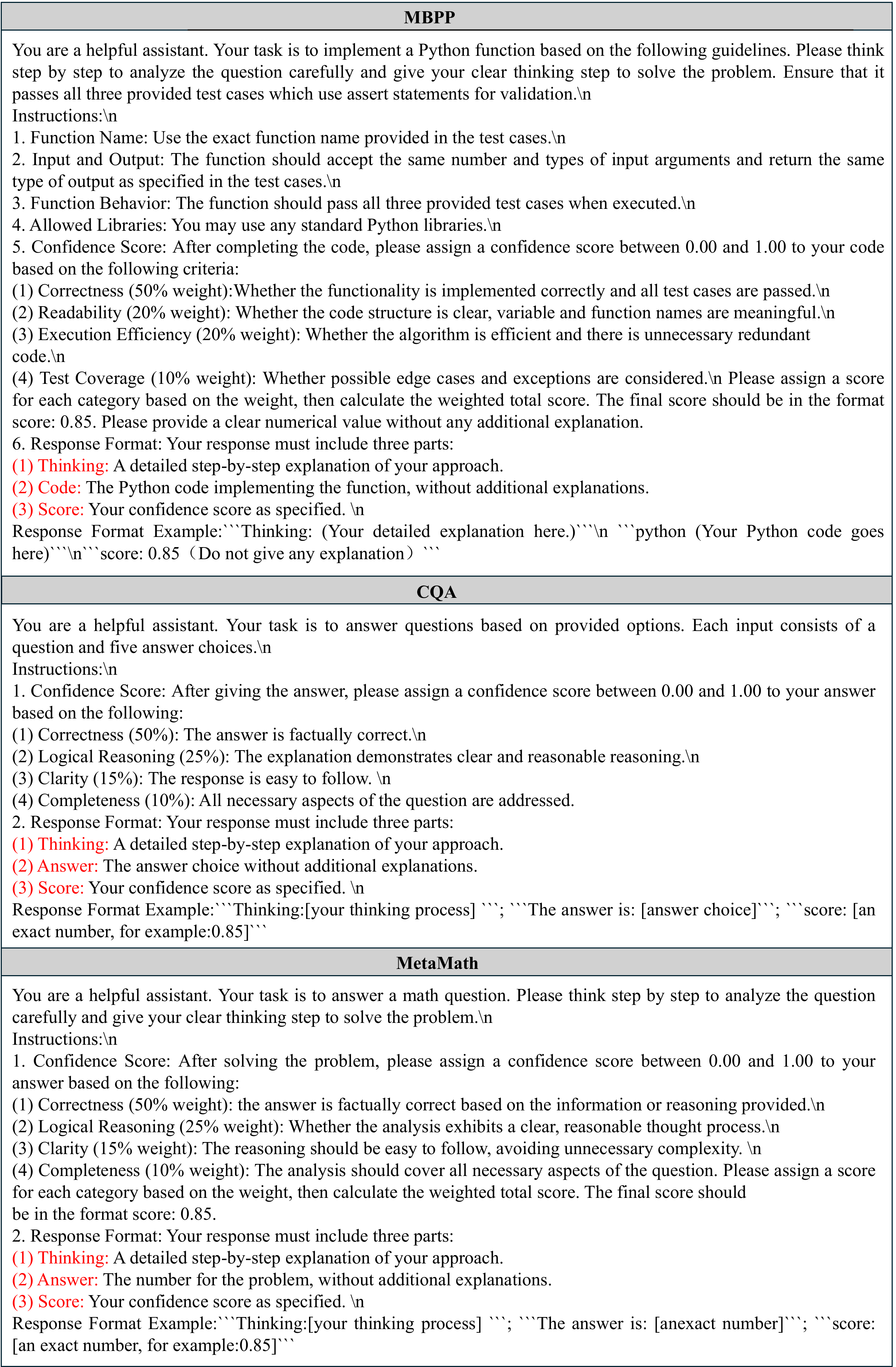} 
	\caption{Prompt for Teachers to Generate Pseudo-Labels on the MBPP Dataset.} 
	\label{fig:prompt_1} 
\end{figure*}

The specific prompt for teachers to generate pseudo-labels on the MBPP, CQA and MetaMath dataset is shown in Figure \ref{fig:prompt_1}.


\section{Case Study on MBPP Dataset}
\label{appendix:case}

\subsection{Query}
\begin{figure*}[t] 
	\centering 
	\includegraphics[width=0.9\textwidth]{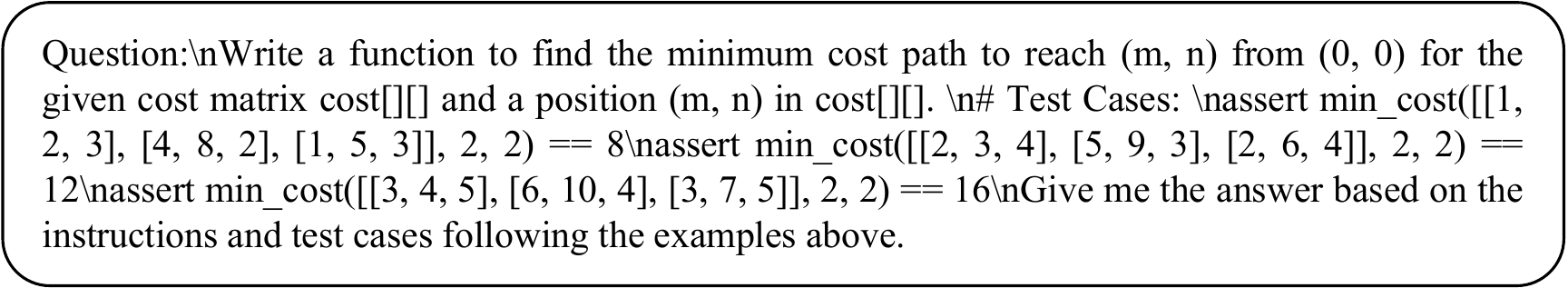} 
	\caption{An example query of the MBPP dataset, showing the input programming task and its corresponding test case.} 
	\label{fig:query} 
\end{figure*}

A specific query example of the MBPP dataset is shown in Figure \ref{fig:query}.

\subsection{Teacher’s Original Response (Dataset of HaRD Stage-1)}
\begin{figure*}[htbp] 
	\centering 
	\includegraphics[width=0.9\textwidth]{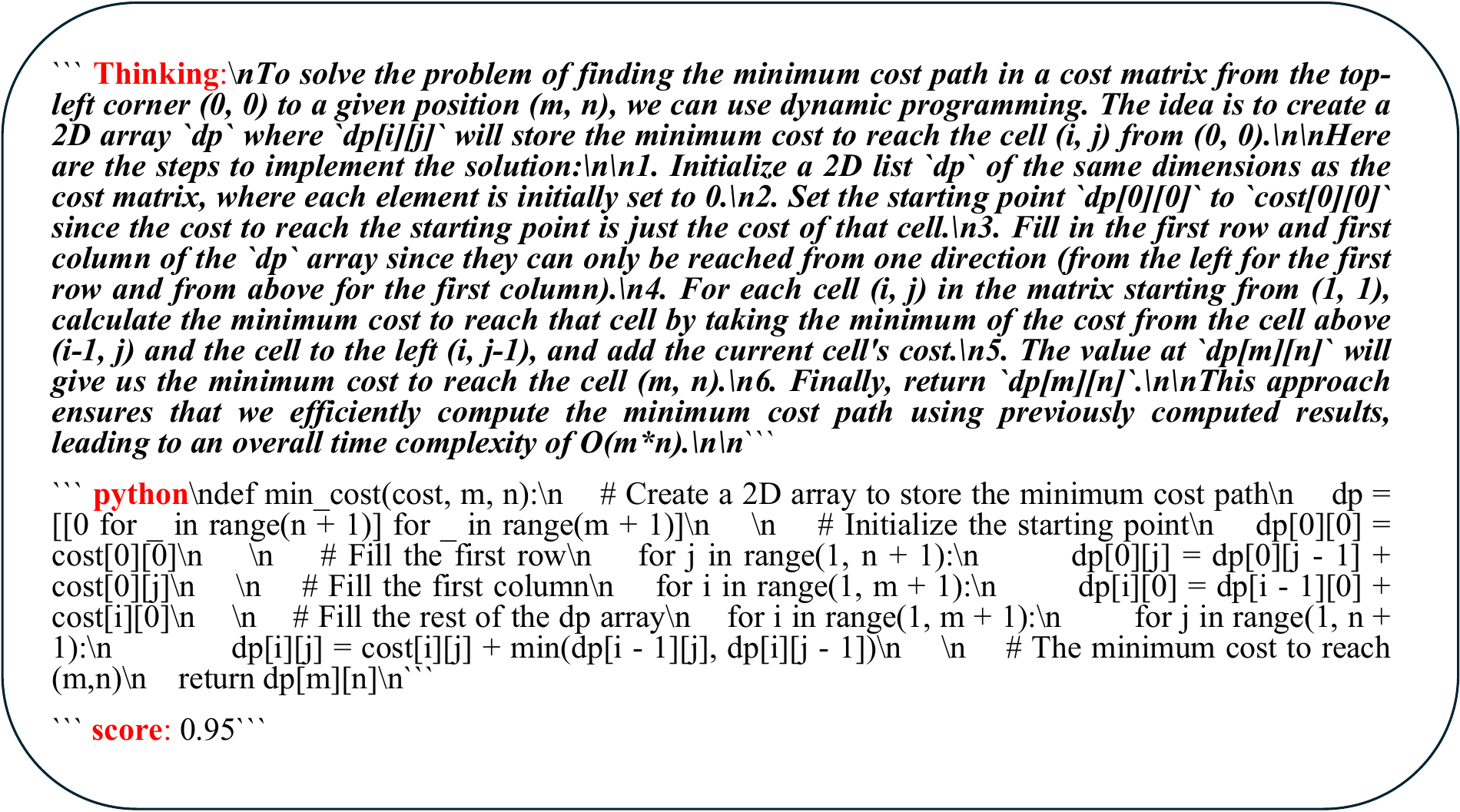} 
	\caption{The original structured response generated by the teacher model based on the query. The italicized part represents the full reasoning process, which will later be refined based on the student model’s response.} 
	\label{fig:Original_Response} 
\end{figure*}

After Dual-Criteria rejection sampling, one of the selected teacher responses (with full reasoning process) is shown in Figure \ref{fig:Original_Response}.

\subsection{Student's Inference for Same Query}
\begin{figure*}[htbp] 
	\centering 
	\includegraphics[width=0.9\textwidth]{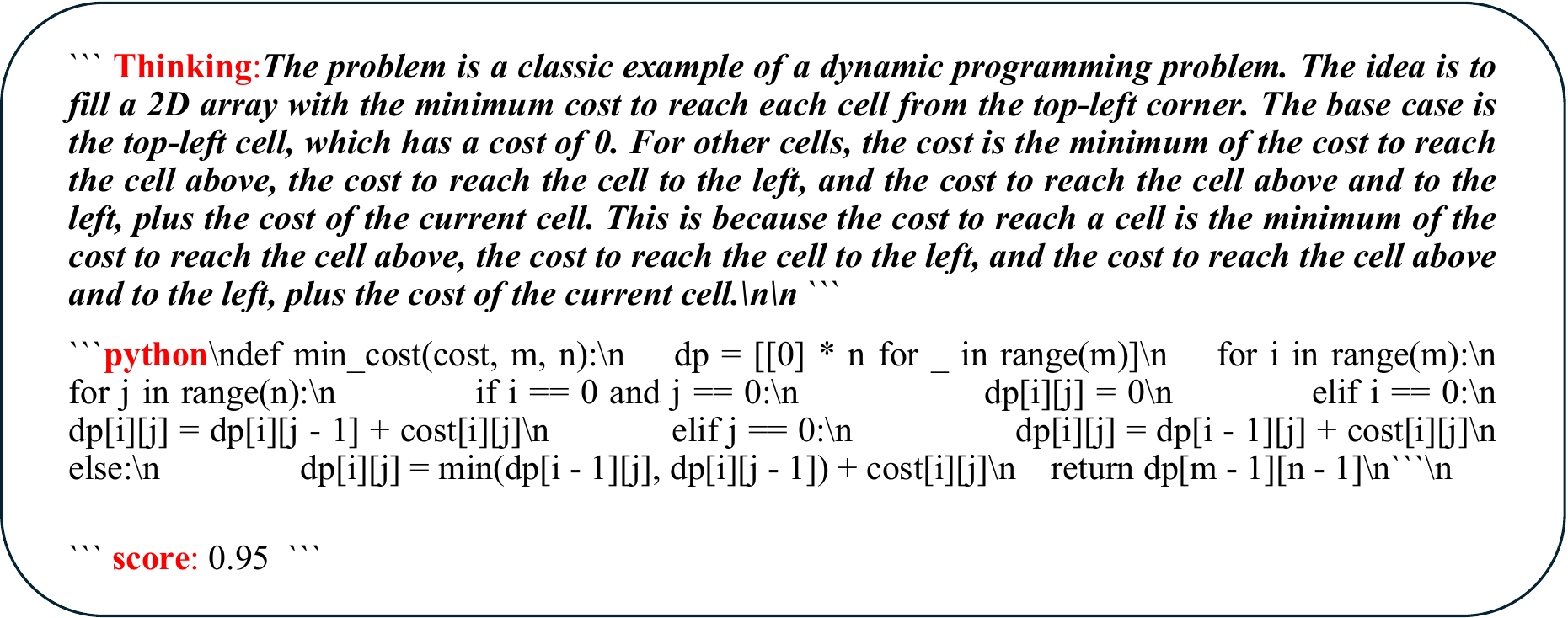} 
	\caption{The inference results of the student model after the first stage of SFT on the distill dataset.} 
	\label{fig:student_inf} 
\end{figure*}

After performing SFT on the student model using the full reasoning dataset from the first stage, an initial student model with reasoning ability is obtained. Let this stage’s student model perform inference on the distillation dataset, resulting in the student’s response, as shown in Figure \ref{fig:student_inf}.

\subsection{Teacher's Refinement
 Prompt}
\begin{figure*}[htbp] 
	\centering 
	\includegraphics[width=0.9\textwidth]{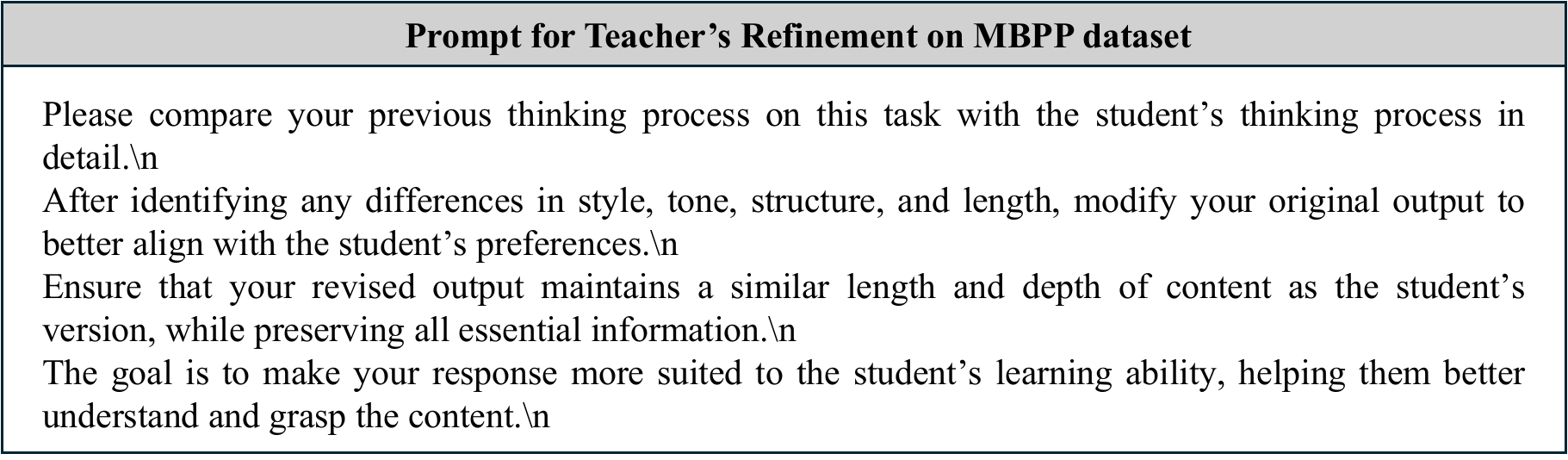} 
	\caption{Prompt for teacher's refinement.} 
	\label{fig:refine} 
\end{figure*}

After the student model performs inference, the reasoning processes of both the student and teacher models are extracted for the same query. The teacher model modifies its original reasoning process based on the student model’s response preferences. The prompt is shown in Figure \ref{fig:refine}.

\subsection{Compressed Reasoning Data after Refinement (Dataset of HaRD Stage-2)}
\begin{figure*}[htbp] 
	\centering 
	\includegraphics[width=0.9\textwidth]{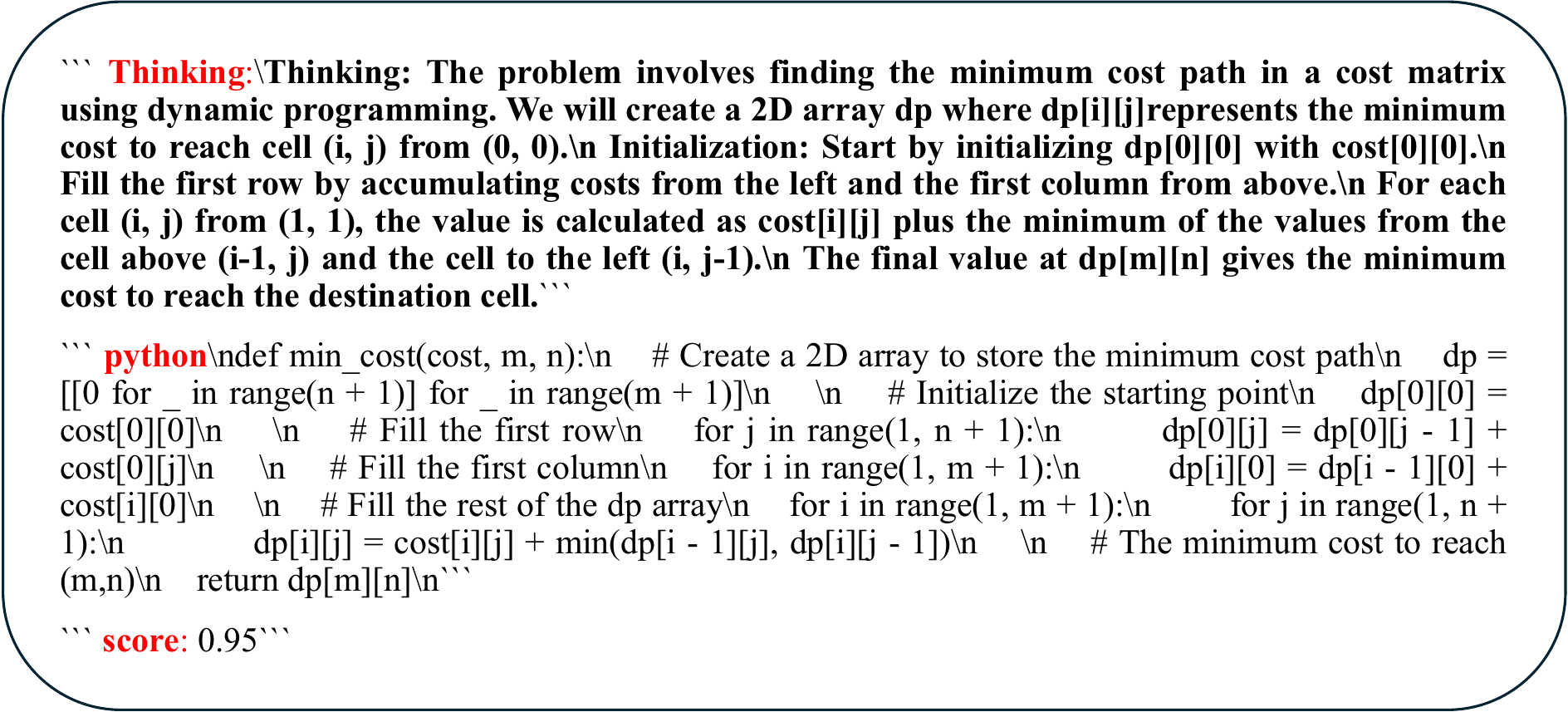} 
	\caption{Dataset of reasoning-compressed 
distillation stage.} 
	\label{fig:reasoning2} 
\end{figure*}

By having the teacher model refine its original response based on the student model’s answer to the same question, the teacher’s response can better align with the student’s learning characteristics. This can be compared to the human learning process: the teacher delivers the initial lesson, then, after receiving feedback from the student, modifies the lesson plan based on the feedback to improve subsequent teaching. The teacher’s revised response is shown in Figure \ref{fig:reasoning2}. Since the refined reasoning is intuitively much shorter than the original reasoning, we refer to it as the “reasoning-compressed dataset”, which is used for the second stage of distillation in HaRD.

\subsection{Reasoning-free
Distillation Data}
After the second stage of SFT, the student model gradually internalizes the reasoning process into its own capability. In the third stage, we aim to enable the student model to learn query-answer end-to-end, strengthening its memory of the answer. Through this repetitive training, the student model will eventually be able to provide accurate answers without outputting reasoning. The format of the distillation data in the third stage is: prompt + query + answer (python code), meaning that the reasoning process is entirely removed. This dataset is used to continue SFT on the student model from the second stage, resulting in the final trained student model, completing the entire HaRD training process.

\section{Details of Confidence Score Selection}
\label{appendix:confidence}

Taking the MBPP dataset as an example, we prompt the LLM to assign a confidence score to its own output based on four weighted criteria: correctness (50\%), readability (20\%), execution efficiency (20\%), and test coverage (10\%). The final score is returned in the format score: 0.85 as a clear numeric value.

Inspired by recent work on LLM self-consistency, which shows that LLMs can assess the quality of their own outputs by assigning confidence scores, we adopt a similar strategy to evaluate the trustworthiness of generated samples. Specifically, we introduce a confidence threshold: samples with scores above this threshold are retained as reliable, while those below are discarded.

\begin{figure*}[htbp] 
	\centering 
\includegraphics[width=0.85\textwidth]{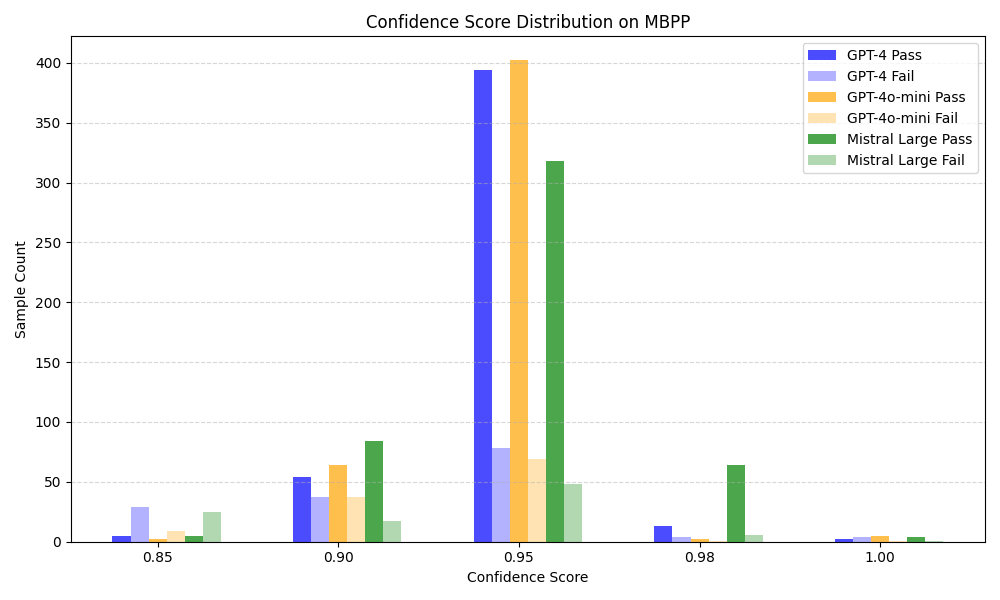} 
	\caption{Confidence Score Distribution on MBPP dataset via GPT-4, GPT-4o-mini and Mistal-Large.} 
	\label{fig:mbpp} 
\end{figure*}

\begin{figure*}[htbp] 
	\centering 
\includegraphics[width=0.85\textwidth]{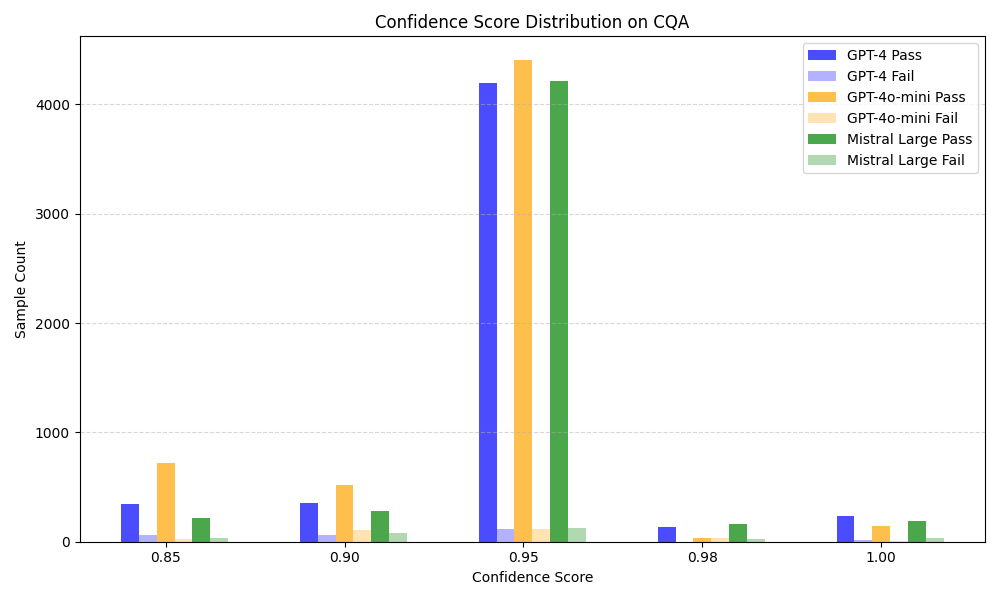}
	\caption{Confidence Score Distribution on CQA dataset via GPT-4, GPT-4o-mini and Mistal-Large.}
	\label{fig:CQA} 
\end{figure*}

\begin{figure*}[htbp] 
	\centering 
\includegraphics[width=0.85\textwidth]{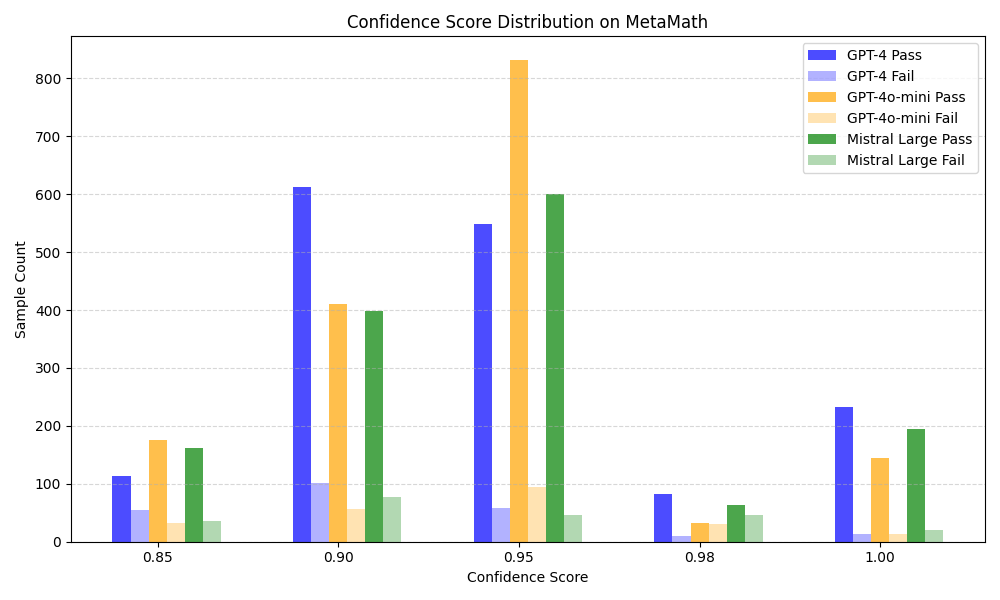} 
	\caption{Confidence Score Distribution on MetaMath dataset via GPT-4, GPT-4o-mini and Mistal-Large.} 
	\label{fig:math} 
\end{figure*}

To determine this threshold, we analyze the relationship between confidence scores and output quality. On datasets such as MBPP, CQA, and MetaMath, we observe that when the confidence score is around 0.95, the alignment between model predictions and ground truth is the highest (e.g., the largest number of passed test cases on MBPP). Based on this observation, we set the confidence threshold to 0.95 in our experiments. The confidence score distributions on different datasets are shown in Figures~\ref{fig:mbpp}--\ref{fig:math}.

 \end{document}